\documentclass{article}

\usepackage{arxiv}

\usepackage[utf8]{inputenc} 
\usepackage[T1]{fontenc}    
\usepackage{hyperref}       
\usepackage{url}            
\usepackage{booktabs}       
\usepackage{amsfonts}       
\usepackage{nicefrac}       
\usepackage{microtype}      
\usepackage{lipsum}		
\usepackage{graphicx}
\usepackage{natbib}
\usepackage{doi}
\usepackage{amsmath,amssymb}
\usepackage{multirow}
\usepackage{adjustbox}

\title{Improving accuracy in short mortality rate series: Exploring Multi-step Forecasting Approaches in Hybrid Systems}

\author{ \href{https://orcid.org/0000-0002-2038-4115}{\includegraphics[scale=0.06]{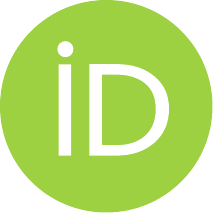}\hspace{1mm}Filipe C. L.~Duarte}\thanks{Corresponding author.} \\
	Departamento de Finanças e Contabilidade\\
	Universidade Federal da Paraíba\\
	João Pessoa, PB-Brazil \\
	\texttt{fcld@academico.ufpb.br} \\
	\And
	\href{https://orcid.org/0000-0002-2396-7973}{\includegraphics[scale=0.06]{orcid.pdf}\hspace{1mm}Paulo S. G.~de Mattos Neto} \\
	Centro de Informática\\
	Universidade Federal de Pernambuco\\
	Recife, PE-Brazil \\
	\texttt{psgmn@cin.ufpe.br} \\
    \AND
    \href{https://orcid.org/0000-0002-3308-2650}
    {\includegraphics[scale=0.06]{orcid.pdf}\hspace{1mm}Paulo R. A.~Firmino} \\
    Departamento de Estatística e Informática \\
    Universidade Federal do Cariri \\
    Juazeiro do norte, CE-Brazil \\
    \texttt{paulo.firmino@ufca.edu.br}
}

\renewcommand{\shorttitle}{\textit{arXiv} Template}

\hypersetup{
pdftitle={A template for the arxiv style},
pdfsubject={q-bio.NC, q-bio.QM},
pdfauthor={David S.~Hippocampus, Elias D.~Striatum},
pdfkeywords={First keyword, Second keyword, More},
}

\begin{document}
\maketitle

\begin{abstract}
    The decline in interest rates and economic stabilization has heightened the importance of accurate mortality rate forecasting, particularly in insurance and pension markets. Multi-step-ahead predictions are crucial for public health, demographic planning, and insurance risk assessments; however, they face challenges when data are limited. Hybrid systems that combine statistical and Machine Learning (ML) models offer a promising solution for handling both linear and nonlinear patterns. This study evaluated the impact of different multi-step forecasting approaches (Recursive, Direct, and Multi-Input Multi-Output) and ML models on the accuracy of hybrid systems. Results from 12 datasets and 21 models show that the selection of both the multi-step approach and the ML model is essential for improving performance, with the ARIMA-LSTM hybrid using a recursive approach outperforming other models in most cases.
\end{abstract}

\keywords{short time series \and hybrid systems \and multi-step forecasting \and mortality rate forecasting.}

\section{Introduction}\label{sec:Intro}
Accurate forecasting of mortality rates has implications for a wide range of sectors, including insurance, pension funds, demography, health, and government.
In this sense, insurers and pension funds model the dynamics of population mortality and try to ensure adequate assessments of long-term liability, pricing strategies, and risk management~\citep{mitchell2013, bravo2018valuation}. 

The existing literature categorizes mortality forecasting approaches into multivariate and univariate~\citep{feng:2018}.
The Age-Period-Cohort (APC) framework, exemplified by Lee-Carter (LC)~\citep{Lee:92} and Plat~\citep{plat2009stochastic}, dominates the multivariate category. Although famous for capturing historical trends, APC faces limitations because of its static representation of age-specific dynamics~\citep{mitchell2013}. Univariate models, by age-specific time series modeling such as Autoregressive Integrated Moving Average (ARIMA)~\citep{mcnown:1989}, Exponential Smoothing (ETS)~\citep{feng:2018}, and Gaussian Process Regression~\citep{wu:18}, offer promising alternatives by modeling individual age groups without demanding cross-correlations. 

Alternatively, Machine Learning (ML) models have emerged as a new wave of mortality forecasting approaches~\citep{bravo2021forecasting}. Their data-driven, trainable, and nonlinear nature holds significant promise \citep{demattos:2017}. However, relying solely on a single ML model may not adequately capture the intricate interplay of the linear and nonlinear patterns inherent in real-world time series~\citep{zhang:2003, demattos:2017, de2021hybrid}. In addition, ML models can be susceptible to underfitting and overfitting problems, which further hampers their performance~\citep{demattos:2017,de2021hybrid}. Therefore, an alternative to solving this problem is the employment of hybrid systems, which model linear and nonlinear time series patterns separately~\citep{zhang:2003}. This class of hybrid systems employs (i) a linear statistical model for time-series forecasting, (ii) an ML model to predict the residual (error) series from the linear estimate, and (iii) the combination (e.g., simple addition) of the two forecasts from (i) and (ii). Thus, the strengths of linear and nonlinear models are used to improve forecast accuracy by combining these techniques~\citep{demattos:2017}.

Nevertheless, the literature ~\citep{khashei:2011, de2021hybrid} indicates that there is no single ML model that produces the best forecasting accuracy across all residual series. This can be explained by the fact that these series are often heteroscedastic and noisy owing to different temporal patterns and random oscillations~\citep{de2021hybrid}. Hence, these traits make model selection, specification, and training difficult ~\citep{de2021hybrid}.
In addition, multi-step forecasting is crucial for stakeholders in mortality rate forecasting, where long-term predictions are crucial. Users require sequences of future data points and demand forecasts for entire periods. In this context, two primary challenges emerge when designing hybrid systems for the multi-step forecasting of mortality rate series: What is the multi-step forecasting approach most appropriate for residual series modeling? Which ML model is the most accurate in error series forecasting?

This study investigates the influence of the multi-step forecasting approach and the choice of the ML model on the accuracy of hybrid systems. This study addresses mortality rate forecasting because of its importance in several application areas and because it deals with short time series. Experiments were conducted to assess three main multi-step approaches (Recursive, Direct, and Multi-Input Multi-Output (MIMO)~\citep{taieb2012review}) employing three ML models, Multilayer Perceptron (MLP)~\citep{zhang:2003}, Long Short-Term Memory (LSTM)~\citep{liu2019cssap}, and N-BEATS~\citep{deng2022ip}, in the residuals forecasting. The nine evaluated hybrid systems were compared with each other and with single models from the literature, encompassing ML, statistical, and multivariate models using 12 datasets. Therefore, the main contributions of this study are as follows: the multi-step recursive approach leads to better accuracy in residual modeling; The LSTM employing a multi-step recursive approach is the most suitable model for residual forecasting among the evaluated ML models; and the hybrid system ARIMA+LSTM attains superior accuracy regarding the Mean Absolute Percentage Error (MAPE) than single ML, statistical, and multivariate models for forecasting mortality rates. 

The main contributions of this study are as follows:

1. The multi-step recursive approach leads to better accuracy in residual modeling for mortality rate forecasting.

2. The LSTM model employing a multi-step recursive approach was found to be the most suitable for residual forecasting among the evaluated machine learning models.

3. The hybrid ARIMA+LSTM system with a recursive multi-step-ahead approach attained superior accuracy in terms of the Mean Absolute Percentage Error (MAPE) compared to single machine learning, statistical, and multivariate models for forecasting mortality rates.

The remainder of this paper is organized as follows.
Section~\ref{sec:Methods} describes the methods used for time-series forecasting. Section~\ref{sec:Exp} presents the experiments. The results and discussion of the experiments are reported in Section~\ref{sec:Res}. The conclusions are presented in Section~\ref{sec:Conc}.

\section{Background}\label{sec:Methods}

This section consists of three subsections that describe the techniques utilized in this study. Section~\ref{sec:Hybrid} outlines the hybrid systems used in time-series forecasting. Section~\ref{sec:multi-step} details the multi-step forecasting approaches. Section~\ref{sec:TimeSeries} briefly introduces the time-series forecasting models.

\subsection{Hybrid systems}\label{sec:Hybrid}

Hybrid systems, which combine linear and nonlinear models, have become prominent owing to their superior accuracy and flexibility in forecasting~\citep{zhang:2003, demattos:2017}. \cite{zhang:2003} introduced the hybrid model foundations from the supposition that a time series $Z_t$ is composed by linear combining of linear ($L_t$) and nonlinear patterns ($N_t$), as shown in Equation \ref{eq1}:
\begin{equation}
Z_t=L_t + N_t \label{eq1}.
\end{equation}

In this system, the first step consists of generating a linear estimate $\hat{L}_t$. Well-consolidated statistical fundamentals \cite{box1970distribution} and Box  and Jenkins methodology \cite{box2015time} guarantee the proper modeling of the linear patterns present in $Z_t$. The residual $E_t$ of the linear estimate $\hat{L}_t$ for $Z_t$ is calculated as shown in Equation \ref{eq2}.
\begin{equation}
E_t = Z_t - \hat{L}_t \label{eq2}.
\end{equation}

Although an ARIMA model can be applied to $Z_t$, its residuals may exhibit nonlinear characteristics according to Zhang's hypothesis described in Equation~\ref{eq1}. To address this,~\cite{zhang:2003} used an MLP to model these residuals, resulting in the nonlinear forecast $\hat{N}_t$. ~\cite{zhang:2003} emphasized the necessity of accurate linear modeling within the time series to ensure the randomness of the residuals and the absence of linear autocorrelations. The final forecast ($\hat{Z}_t$) of the hybrid model is the sum of the linear ($\hat{L}_t$) and nonlinear forecasts ($\hat{N}_t$), as shown in Equation \ref{eq:forecast_hybrid}.
\! \! \! \!
\begin{equation}
\hat{Z}_t = \hat{L}_t + \hat{N}_t \label{eq:forecast_hybrid}.
\end{equation}
\! \! \! \!
Initially focused on one-step-ahead predictions, Zhang's model~\citep{zhang:2003} has since been expanded to various applications, including stock market~\citep{pai_lin:2005}, wind speed~\citep{do2018innovative}, and epidemic~\citep{chakraborty2019forecasting}. 
However, it is essential to note that most studies with hybrid systems evaluate one-step forecasting, as highlighted in studies such as those by~\cite{shi2012evaluation}. 
Recently,~\cite{duarte2024hybrid} proposed a hybrid system for multi-step mortality rate forecasting of the French population that combined the recursive and direct approaches and outperformed single statistical and ML models.

\subsection{Multi-step forecasting approaches}\label{sec:multi-step}

Multi-step-ahead forecasting in time series is a complex and challenging task that presents different challenges compared to one-step forecasting, such as error accumulation and increased uncertainty on longer prediction horizons~\citep{hamzaccebi2009comparison, taieb2012review}. In this context, given a time series ${z_1, z_2, ..., z_t}$ comprising observations $t$, the objective is to predict the next $H$ future observations ${z_{t+1}, z_{t+1}, ..., z_{t+H}}$. 

Traditional approaches to multi-step forecasting include Recursive, Direct, and MIMO~\citep{KALATEAHANI2019689, taieb2012review}. The Recursive approach involves training a single model $f$ to forecast one step at a time \citep{taieb2012review},
\begin{equation}
z_{t+1} = f(z_t, ...,z_{t-d+1}) + \epsilon,
\end{equation}
where $d$ represents the number of time lags.

To forecast $H$ horizons, initially, $f$ forecasts the first horizon $z_{t+1}$. This forecast is then used as an input to predict the second horizon using the same model $f$. This process continues until predictions are obtained for all desired horizons. Mathematically, with a trained model $\hat{f}$, the forecasts are expressed as 
\begin{equation}
\resizebox{0.70\textwidth}{!}{$
\hat{z}_{N+h} = \begin{cases}
\hat{f}(z_N,...,z_{N+d-1}) & \text{if } h = 1 \\
\hat{f}(\hat{z}_{N+h-1},...,\hat{z}_{N+1},z_N,...,z_{N-d+h}) & \text{if } h \in \{2,...,d\} \\
\hat{f}(\hat{z}_{N+h-1},...,\hat{z}_{N+h-d}) & \text{if } h \in \{d+1, ..., H\} \\
\end{cases}$}
\end{equation}

The Direct approach involves performing independent forecasts for each horizon \citep{hamzaccebi2009comparison, taieb2012review}. A specific model $f_h$ is generated for each future horizon independently,
\begin{equation}
z_{t+h} = f_h(z_t, ...,z_{t-d+1}) + \epsilon,
\end{equation}
where $t \in \{d,...,N-H\}$.
Forecasts are obtained using $H$ trained models $\hat{f_h}$,
\begin{equation}
\hat{z}_{N+h} = \hat{f}_h(z_N,...,z_{N-d+1}).
\end{equation}

The MIMO approach predicts multiple outputs simultaneously \citep{taieb2012review, bontempi:2008}. It was developed to maintain the stochastic dependency associated with multiple forecasts \citep{bontempi:2008, taieb2012review}, which was previously overlooked by the other two strategies.
In MIMO, a model $F$ learns the association between input data and $H$ forecast horizons in one go,
\begin{equation}
[z_{t+H}, ..., z_{t+1}] = F(z_t,...,z_{t+d-1}) + \boldsymbol{\epsilon},
\end{equation}
where $t \in {d, ..., N-H}$ and $F: \mathbb{R}^d \Rightarrow \mathbb{R}^H$ is a vector function, with $\boldsymbol{\epsilon}$ being a white noise vector with a non-diagonal covariance matrix \cite{taieb2012review}.
Forecasts are obtained all at once through a multi-output model $\hat{F}$,
\begin{equation}
[\hat{z}_{t+H}, ..., \hat{z}_{t+1}] = \hat{F}(z_N,...,z_{N-d+1})
\end{equation}

Although the Recursive approach is computationally less demanding and beneficial for short series, it may be prone to error accumulation~\citep{ming2014multistep, taieb2012review}. In contrast, the direct method avoids such accumulation but overlooks temporal interdependencies~\citep{taieb2009long}. The MIMO approach attempts to address these shortcomings. However, it confines all forecast horizons to a single model in a multi-output way~\citep{taieb2009long} and, therefore, ends up requiring more data than the recursive approach. These complexities highlight the need to choose the most appropriate multi-step forecasting approach for each problem~\citep{taieb2009long, taieb2012review, ouyang2022deep}.

\subsection{Time series forecasting models}\label{sec:TimeSeries}
Time series forecasting is typically approached through the application of linear and nonlinear models or a combination of both~\citep{zhang:2003, demattos:2017}. 
The ARIMA model is the most widely used and well-established model in the literature on linear modeling~\citep{zhang:2003, demattos:2017, de2021hybrid}. Artificial neural networks are commonly chosen for nonlinear modeling~\citep{zhang:2003,demattos:2017,de2021hybrid}. Concerning the combination, hybrid systems based on the decomposition of a time series into linear and nonlinear components have demonstrated relevant results in terms of accuracy and commonly outperform the single-model approach~\citep{zhang:2003, demattos:2017, chakraborty2019forecasting}.

ARIMA is a statistical linear model that combines two models (i.e., Autoregressive (AR) and Moving Average (MA)) and an integration operator. The AR part models the pattern of the time series about past lags, while the MA models the errors resulting from the model to the past values of the time series. The integrated part of the model performs the differencing step to transform the time series into stationary data ~\citep{box2015time}.

In terms of ML models, MLP, Recurrent Neural Networks (RNN), such as LSTM, and N-BEATS are promising alternatives commonly used in various areas of knowledge to forecast time series~\citep{oreshkin2019n,chandra:2021}. 
The MLP is a feedforward neural network with multiple neurons distributed in parallel with a layered composition, including an input layer, one or more hidden layers, and an output layer~\citep{demattos:2017}. MLP are commonly used in forecasting literature~\citep{zhang:2003, hamzaccebi2009comparison, chakraborty2019forecasting}.
The LSTM model effectively overcomes the challenges of traditional RNNs in managing long-term dependencies, showing promising results in forecasting time series in various applications~\citep{chandra:2021,lara2021experimental}. 
The architecture of the LSTM network is characterized such that each unit has a cell state and three regulatory ports: input, output, and forget. These gates control the flow of information, enabling the network to maintain or ignore patterns in long sequences, which is a crucial feature for time-series forecasting~\citep{chandra:2021}. As LSTM processes time-series data, each unit dynamically updates its internal state by integrating the current input, preceding state, and output gates~\citep{chandra:2021}. 
The N-BEATS adopts a fully neural network-based approach, where the model comprises stacks of fully connected layers with forward and backward residual links~\citep{oreshkin2019n}. Each stack, termed a ``block,” is designed to forecast time series directly, and the model employs a basis expansion strategy for its outputs~\citep{oreshkin2019n}. In particular, N-BEATS shows promising flexibility and scalability, which makes it adept at handling various time-series forecasting tasks simultaneously (that is, global forecasting) without the need for task-specific customization (fine-tuning)~\citep{oreshkin2019n}. 

\section{Experiments}\label{sec:Exp}
This Section outlines the experiments.
First, Section~\ref{sec:DataSetup} presents the datasets and data partitions used in this study. Then, Section~\ref{sec:Settings} provides the description of the experimental setup, which elucidates the methodology employed to address the problems pointed out in Section~\ref{sec:Intro}.

\subsection{Dataset and setup}\label{sec:DataSetup}
The experiment used annual crude mortality rate data from four countries: Australia, France, Japan, and Portugal. Data were retrieved from the Human Mortality Database (HMD)~\citep{HMD}. These time series were modeled specifically by age, comprising ages between 0 and 100 years and covering the years 1950-2019 (1921-2019 for Australia). Before 1950, data for Europe and Japan were excluded because of potential confounders, such as World Wars and pandemics, which significantly impacted historical mortality patterns. Following established practices~\citep{Lee:92, wu:18}, natural logarithm transformation was applied to the data to reduce skewness. Data partitioning followed previous work~\citep{wu:18}, with the first sample (1921/1950–2009) serving as a training set and the second (2010-2019) serving as a test set. For the machine learning models, further splitting occurred within the training set, reserving 80\% for training and 20\% for hyperparameter validation. Both the time and residual series were normalized to the [0, 1] interval for all ML experiments.

\subsection{Experimental settings}\label{sec:Settings}

For all models, except for the multivariate mortality models (i.e., LC and Plat) that used all age groups, the complete logarithmic mortality curve forecasts were generated for the validation and test sets using the specific ages: 0, 1, 2, 5, 10, 12, 15, 18, 20, 22, 25, 28, 30, 40, 50, 60, 70, 80, 90, and 100. Cubic spline interpolation based on Wu and Wang’s methodology~\citep{wu:18} facilitated this process.

The hybrid system employs independent linear and nonlinear models. ARIMA, known for its simplicity and versatility, was chosen for linear modeling~\citep{zhang:2003, chakraborty2019forecasting, de2021hybrid, duarte2024hybrid}. Three widely used time-series forecasting ML models were utilized: LSTM, MLP, and N-BEATS~\citep{oreshkin2019n, de2021hybrid, duarte2024hybrid}.

Comparisons were made with existing models to assess the performance of the hybrid system. These included ARIMA, LC~\citep{Lee:92}, PLAT~\citep{plat2009stochastic}, hybrid systems such as ARIMA-MLP, ARIMA-LSTM, and ARIMA-N-BEATS (with Recursive, MIMO and Direct approaches) adapted from~\citep{zhang:2003}, six single machine learning models (MLP, LSTM, and N-BEATS) using Recursive~\citep{nigri2019deep}, Direct~\citep{livieris2022novel, duarte2024hybrid}, and MIMO~\citep{bontempi:2008, chandra:2021, livieris2022novel} approaches. The root-mean-square error (RMSE) was the primary metric used for hyperparameter validation. 
Bayesian optimization, recognized for its effectiveness in computationally expensive scenarios~\citep{frazier2018}, was used to optimize the hyperparameters and select the optimal configuration for each nonlinear model within the search space defined in Table~\ref{tabGrid}. This is consistent with the findings of recent studies on hybrid systems [5, 6]. Importantly, ``Units in the hidden layer'' and ``Learning rate'' represent closed intervals sampled on a logarithmic scale within the specified ranges. For each ML model, ten runs were performed, each representing a unique combination of hyperparameters, and five independent runs were executed. 

\begin{table}[h!]
\caption{Hyper-parameters values used in the Bayesian optimization.}
\label{tabGrid}
\centering
\scalebox{0.85}{
\begin{tabular}{l|l|l}
\hline
Model                 & Parameter             & Possible values                                  \\ \hline
\multirow{6}{*}{MLP}                  & Maximum number of iterations  & 500                                   \\
                      & Units in the input layer  & 2                             \\
                      & Units in the hidden layer & {[}2, 100{]}    \\
                  & Activation function   & Hyperbolic tangent, ReLU               \\
                      & Optimizer             & Adam                     \\
                      & Learning rate         &  [1E-4, 1E-1]                                 \\
                       \hline
 \multirow{5}{*}{LSTM}                 & Maximum number of iterations  & 500                                  
\\
                      & Units in the input layer  & 2                           \\
                      & Units in the hidden layer & {[}2, 100{]} \\
                      & Optimizer             & Adam                                   \\ 
                      & Learning rate         &  [1E-4, 1E-1]      \\
\hline                      
    \multirow{8}{*}{N-BEATS}                    & Maximum number of iterations & 500 \\
                        & Units in the input layer & 2 \\
                        & Number of blocks per stack & 4 \\
                        & Number of hidden layers & [1, 2, 3, 4] \\ 
                        & Stack type & Generic \\
                        & Units in the hidden layer & {[2, 100]} \\
                        & Optimizer & Adam \\
                        & Learning rate &  [1E-4, 1E-1]   \\
\hline
\end{tabular}
}
\end{table}

MAPE was used as a metric to assess the accuracy of the model forecast. This choice aligns with established practices in time-series forecasting, particularly when comparing forecast accuracy across datasets~\citep{garcia2015electricity}. The essential advantage of the MAPE lies in its scale invariance, enabling direct comparison of performance across diverse datasets with varying scales. The forecast accuracy of the hybrid system was evaluated using a three-stage process. In Stage 1, the focus was on the multi-step-ahead approach within each hybrid model, comparing the performance of the Direct, MIMO, and Recursive approaches. This stage aimed to identify the most effective multi-step-ahead approach for capturing the residual dynamics. In stage 2, we used the optimal multi-step approach identified in stage 1 to compare the overall forecast performance of complete hybrid systems. In the final stage of the evaluation, we compared the best hybrid system with alternatives using MAPE and Percentage Difference (PD) as evaluative metrics. The PD is denoted in Equation~\ref{eq:PD}, 
\begin{equation}
PD = 100\cdot\frac{Error_{a}-Error_{r}}{Error_{a}},
\label{eq:PD}
\end{equation}
where $Error_{r}$ is the MAPE value of the reference, and $Error_{a}$ is the MAPE value of the alternative.

\section{Results and discussion}\label{sec:Res}

Section~\ref{sec:Resmulti-step} compares multi-step forecasting approaches for residual series modeling, Section~\ref{sec:ResModel} discusses the accuracy of ML models in forecasting residual series, and Section~\ref{sec:Resliterature} presents a comparative analysis between the best hybrid model and literature approaches.

\subsection{Analyzing the performance of the multi-step approaches}\label{sec:Resmulti-step}

Table~\ref{tab:mape_approaches} shows the MAPE value for the test set for the 12 datasets analyzed with the hybrid systems ARIMA-LSTM, ARIMA-MLP, and ARIMA-N-BEATS employing Direct, MIMO and Recursive approaches. For the ARIMA-LSTM model, the recursive approach exhibited superior accuracy over the Direct and MIMO approaches in most cases, specifically in seven of the 12 datasets. In the context of the ARIMA-MLP model, the recursive approach consistently outperformed its counterparts, ensuring the top position in forecasting accuracy across two-thirds of the datasets, totaling eight out of twelve. Concerning the ARIMA-N-BEATS model, the Recursive approach performs better than the others, reaching the smallest MAPE in 6 of 12 datasets, particularly for the female population across all datasets except France.

\begin{table}[h!]
    \caption{MAPE (\%) values in the test set obtained by the hybrid systems. The best for each hybrid system and data set is highlighted in bold.}
\centering.
\scalebox{0.95}{
 \begin{adjustbox}{max width=\textwidth}
 \begin{tabular}{lc|ccc|ccc|ccc}
\hline
\multicolumn{2}{c|}{\multirow{2}{*}{Dataset}}            & \multicolumn{3}{c|}{ARIMA-LSTM}                                                         & \multicolumn{3}{c|}{ARIMA-MLP}                                                          & \multicolumn{3}{c}{ARIMA-N-BEATS}                                                        \\ \cline{3-11} 
\multicolumn{2}{l|}{}                                    & \multicolumn{1}{c|}{Direct}        & \multicolumn{1}{c|}{MIMO}          & Recursive     & \multicolumn{1}{c|}{Direct}        & \multicolumn{1}{c|}{MIMO}          & Recursive     & \multicolumn{1}{c|}{Direct}        & \multicolumn{1}{c|}{MIMO}          & Recursive     \\ \hline
\multicolumn{1}{l|}{\multirow{3}{*}{Australia}} & Female & \multicolumn{1}{c|}{2.154}          & \multicolumn{1}{c|}{2.101}          & \textbf{2.072} & \multicolumn{1}{c|}{2.141}          & \multicolumn{1}{c|}{\textbf{2.033}} & 2.185          & \multicolumn{1}{c|}{2.420}          & \multicolumn{1}{c|}{2.185}          & \textbf{2.042} \\ \cline{2-11} 
\multicolumn{1}{l|}{}                           & Male   & \multicolumn{1}{c|}{2.029}          & \multicolumn{1}{c|}{\textbf{1.970}} & 1.988          & \multicolumn{1}{c|}{1.974}          & \multicolumn{1}{c|}{\textbf{1.923}} & 1.982          & \multicolumn{1}{c|}{2.428}          & \multicolumn{1}{c|}{2.391}          & \textbf{2.197} \\ \cline{2-11} 
\multicolumn{1}{l|}{}                           & Total  & \multicolumn{1}{c|}{1.495}          & \multicolumn{1}{c|}{\textbf{1.469}} & 1.483          & \multicolumn{1}{c|}{1.623}          & \multicolumn{1}{c|}{1.497}          & \textbf{1.470} & \multicolumn{1}{c|}{\textbf{1.519}} & \multicolumn{1}{c|}{1.521} & 1.607          \\ \hline
\multicolumn{1}{l|}{\multirow{3}{*}{France}}    & Female & \multicolumn{1}{c|}{1.574}          & \multicolumn{1}{c|}{\textbf{1.506}} & 1.607          & \multicolumn{1}{c|}{1.628}          & \multicolumn{1}{c|}{1.552}          & \textbf{1.525} & \multicolumn{1}{c|}{1.565}          & \multicolumn{1}{c|}{\textbf{1.497}} & 1.502 \\ \cline{2-11} 
\multicolumn{1}{l|}{}                           & Male   & \multicolumn{1}{c|}{2.251}          & \multicolumn{1}{c|}{2.283}          & \textbf{2.220}& \multicolumn{1}{c|}{2.277}          & \multicolumn{1}{c|}{\textbf{2.172}} & 2.418          & \multicolumn{1}{c|}{\textbf{2.115}} & \multicolumn{1}{c|}{2.170}          & 2.222          \\ \cline{2-11} 
\multicolumn{1}{l|}{}                           & Total  & \multicolumn{1}{c|}{\textbf{1.322}} & \multicolumn{1}{c|}{1.680}          & 1.346          & \multicolumn{1}{c|}{\textbf{1.360}} & \multicolumn{1}{c|}{1.679}          & 1.398          & \multicolumn{1}{c|}{\textbf{1.368}} & \multicolumn{1}{c|}{1.516}          & 1.508          \\ \hline
\multicolumn{1}{l|}{\multirow{3}{*}{Japan}}     & Female & \multicolumn{1}{c|}{2.103}          & \multicolumn{1}{c|}{2.006}          & \textbf{1.920} & \multicolumn{1}{c|}{2.104}          & \multicolumn{1}{c|}{2.033}          & \textbf{1.949} & \multicolumn{1}{c|}{2.116}          & \multicolumn{1}{c|}{2.594}          & \textbf{1.998} \\ \cline{2-11} 
\multicolumn{1}{l|}{}                           & Male   & \multicolumn{1}{c|}{2.095}          & \multicolumn{1}{c|}{2.039}          & \textbf{2.026} & \multicolumn{1}{c|}{2.175}          & \multicolumn{1}{c|}{2.042}          & \textbf{2.034} & \multicolumn{1}{c|}{\textbf{1.686}} & \multicolumn{1}{c|}{1.786}          & 2.236          \\ \cline{2-11} 
\multicolumn{1}{l|}{}                           & Total  & \multicolumn{1}{c|}{1.772}          & \multicolumn{1}{c|}{1.700}          & \textbf{1.575} & \multicolumn{1}{c|}{1.720}          & \multicolumn{1}{c|}{1.703}          & \textbf{1.474} & \multicolumn{1}{c|}{2.122}          & \multicolumn{1}{c|}{2.113}          & \textbf{1.629} \\ \hline
\multicolumn{1}{l|}{\multirow{3}{*}{Portugal}}  & Female & \multicolumn{1}{c|}{3.025}          & \multicolumn{1}{c|}{\textbf{3.018}} & 3.125          & \multicolumn{1}{c|}{3.061}          & \multicolumn{1}{c|}{3.140}          & \textbf{2.850} & \multicolumn{1}{c|}{3.003}          & \multicolumn{1}{c|}{3.005}          & \textbf{2.851} \\ \cline{2-11} 
\multicolumn{1}{l|}{}                           & Male   & \multicolumn{1}{c|}{3.144}          & \multicolumn{1}{c|}{3.328}          & \textbf{3.038} & \multicolumn{1}{c|}{3.339}          & \multicolumn{1}{c|}{3.294}          & \textbf{3.149} & \multicolumn{1}{c|}{3.212}          & \multicolumn{1}{c|}{3.269}          & \textbf{2.997} \\ \cline{2-11} 
\multicolumn{1}{l|}{}                           & Total  & \multicolumn{1}{c|}{2.292}          & \multicolumn{1}{c|}{2.221}          & \textbf{2.134} & \multicolumn{1}{c|}{2.344}          & \multicolumn{1}{c|}{2.319}          & \textbf{2.299} & \multicolumn{1}{c|}{2.447}          & \multicolumn{1}{c|}{\textbf{2.317}} & 2.597          \\ \hline
\end{tabular}
\end{adjustbox}
}
\label{tab:mape_approaches}
\end{table}

In summary, ARIMA-LSTM with the Recursive approach achieved the best performance in 58.33\% of the cases, while the MIMO approach reached the smallest MAPE in 33.33\% of the series, and the direct approach attained only 8.33\% of the sets. ARIMA-MLP with the Recursive approach attained the best accuracy in 66.67\% of the series. This frequency is more than double that of the MIMO approach (25.00\%) and surpasses the performance of the direct approach (8.33\%). ARIMA-N-BEATS with the Recursive approach achieves a frequency of 50.00\%, again outperforming the direct (33.33\%) and MIMO (16.67\%) approaches. 

The Direct and MIMO approaches, characterized by their complexity and numerous parameters, require significant data to capture temporal dynamics accurately~\citep{KALATEAHANI2019689}.
These observations suggest that selecting the most appropriate multi-step forecasting approach is crucial in situations with short time-series data. In this context, the inherent simplicity of the recursive approach, combined with its ability to utilize a larger training dataset, provides it with a noticeable advantage, particularly when analyzing small residual series and data with a low signal-to-noise ratio. 

\subsection{Analyzing the performance of ML models in residual series forecasting}\label{sec:ResModel}

Table~\ref{tab:mape_nonlinear_model} shows the MAPE for the Recursive multi-step hybrid models, presenting the performance of the three models used in the residual modeling, LSTM, MLP, and N-BEATS, across 12 datasets. In addition, Table~\ref{tab:mape_nonlinear_model} records the frequency of times each model attained the best result within each dataset, with the best performances highlighted in bold.
The data show that the LSTM model achieved the best performance in five of the 12 cases, equating to a frequency of 41.67\%. This suggests a consistent adaptability of the LSTM model in capturing the temporal dynamics and nonlinear patterns that are inherent in short residual series. Notably, the LSTM model excels in the France dataset with respect to the male and total populations, the female and total populations of Japan, and the total population of Portugal. Although the MLP model performed best in four of the 12 cases (33.33\%), it was particularly effective in the Australian population for both women and men, the total population of Japan, and the Portuguese dataset for women only. The N-BEATS model, with the best result in three of 12 cases (25\%), shows its strengths for females from Australia and France and Portuguese males.

\begin{table}[ht]
\caption{MAPE (\%) in the test set for the Recursive multi-step hybrid systems by each nonlinear model and the frequency of each nonlinear model with a Recursive approach within the hybrid model attained the best performance for all datasets. The performance of the best is in bold.}
\centering
\scalebox{0.9}{
\begin{tabular}{lc|c|c|c|c}
\hline
\multicolumn{2}{c|}{Dataset}        & LSTM            & MLP             & N-BEATS         & Best                                  \\ \hline
\multirow{3}{*}{Australia} & Female & 2.072          & 2.185          & \textbf{2.042} & N-BEATS                               \\
                           & Male   & 1.988          & \textbf{1.982} & 2.197          & MLP                                   \\
                           & Total  & 1.483          & \textbf{1.470} & 1.607          & MLP                                   \\ \hline
\multirow{3}{*}{France}    & Female & 1.607          & 1.525          & \textbf{1.502} & N-BEATS                               \\
                           & Male   & \textbf{2.220} & 2.418          & 2.222 & LSTM                                  \\
                           & Total  & \textbf{1.346} & 1.398          & 1.508          & LSTM                                  \\ \hline
\multirow{3}{*}{Japan}     & Female & \textbf{1.920} & 1.949          & 1.998          & LSTM                                  \\
                           & Male   & \textbf{2.026} & 2.034 & 2.236          & LSTM                                  \\
                           & Total  & 1.575          & \textbf{1.474} & 1.629          & MLP                                   \\ \hline
\multirow{3}{*}{Portugal}  & Female & 3.125          & \textbf{2.850} & 2.851 & MLP                                   \\
                           & Male   & 3.038          & 3.149          & \textbf{2.997} & N-BEATS                               \\
                           & Total  & \textbf{2.134} & 2.299          & 2.597          & LSTM                                  \\ 
                           \hline
\multicolumn{2}{c|}{Frequency (\%)} & \textbf{41.67} & 33.33 & 25.00 \\ \hline                             
\end{tabular}
}
\label{tab:mape_nonlinear_model}
\end{table}

Based on these results, the superiority of LSTM over MLP and N-BEATS can be explained by the fact that the former has the following characteristics: learning long-term dependencies, handling noise, and flexibility. The LSTM model is designed to remember long-term dependencies by leveraging its internal state and gates that control the flow of information.
In this sense, LSTM can learn sequential patterns more effectively than MLP and selectively ignore irrelevant data, whereas MLPs are susceptible to misinterpreting noise as significant signals owing to their static architecture~\citep{chandra:2021, lara2021experimental}. Compared to N-BEATS with its fixed basis expansion for trend and/or seasonality decomposition~\citep{oreshkin2019n}, LSTM offers greater versatility in handling various types of data while mitigating the influence of noise and outliers.

\subsection{Comparing the best hybrid model with literature}\label{sec:Resliterature}

Tables ~\ref{tab:all_models} and ~\ref{tab:mean_rank_std_all} present the forecasting performance in terms of the MAPE, mean, mean rank, and standard deviation of the hybrid ARIMA-LSTM model with a recursive approach against a spectrum of multivariate, statistical, and machine learning models. ARIMA-LSTM shows the best forecast accuracy by the MAPE metric in nine out of 12 datasets, while ARIMA is the second best with three out of 12. For Australia, ARIMA-LSTM shows the lowest MAPE, indicating a pronounced strength in capturing the dynamics of the mortality rate within this country.

\begin{table}[h!]
\centering
\caption{MAPE (\%) of the hybrid system arima-lstm with the Recursive approach compared to multivariate, statistical, and machine learning models. Best results in bold.}
 \begin{adjustbox}{max width=\textwidth}
\begin{tabular}{lc|cc|cccccccccc|c}
\hline
\multicolumn{2}{c|}{\multirow{3}{*}{Dataset}}                           & \multicolumn{2}{c|}{Multivariate}          & \multicolumn{10}{c|}{Single}                                                                                                                                                                                                                                                                       & Hybrid         \\ \cline{3-15} 
\multicolumn{2}{l|}{}                                                   & \multicolumn{1}{c|}{LC}        & PLAT      & \multicolumn{1}{c|}{ARIMA}          & \multicolumn{3}{c|}{LSTM}                                                                 & \multicolumn{3}{c|}{MLP}                                                                  & \multicolumn{3}{c|}{N-BEATS}                                          & ARIMA-LSTM     \\ \cline{3-15} 
\multicolumn{2}{l|}{}                                                   & \multicolumn{1}{c|}{Recursive} & Recursive & \multicolumn{1}{c|}{Recursive}      & \multicolumn{1}{c|}{Direct} & \multicolumn{1}{c|}{MIMO}  & \multicolumn{1}{c|}{Recursive} & \multicolumn{1}{c|}{Direct} & \multicolumn{1}{c|}{MIMO}  & \multicolumn{1}{c|}{Recursive} & \multicolumn{1}{c|}{Direct} & \multicolumn{1}{c|}{MIMO}  & Recursive & Recursive      \\ \hline
\multicolumn{1}{l|}{\multirow{3}{*}{Australia}} & Female                & \multicolumn{1}{c|}{3.823}     & 3.676     & \multicolumn{1}{c|}{2.166}          & \multicolumn{1}{c|}{4.614}  & \multicolumn{1}{c|}{4.870} & \multicolumn{1}{c|}{4.647}     & \multicolumn{1}{c|}{3.506}  & \multicolumn{1}{c|}{3.193} & \multicolumn{1}{c|}{6.417}     & \multicolumn{1}{c|}{3.431}  & \multicolumn{1}{c|}{5.495} & 31.597    & \textbf{2.072} \\ \cline{2-15} 
\multicolumn{1}{l|}{}                           & Male                  & \multicolumn{1}{c|}{4.640}     & 5.385     & \multicolumn{1}{c|}{2.376}          & \multicolumn{1}{c|}{6.170}  & \multicolumn{1}{c|}{6.145} & \multicolumn{1}{c|}{6.323}     & \multicolumn{1}{c|}{4.602}  & \multicolumn{1}{c|}{3.493} & \multicolumn{1}{c|}{5.595}     & \multicolumn{1}{c|}{5.345}  & \multicolumn{1}{c|}{7.434} & 5.988     & \textbf{1.988} \\ \cline{2-15} 
\multicolumn{1}{l|}{}                           & Total                 & \multicolumn{1}{c|}{3.594}     & 4.286     & \multicolumn{1}{c|}{1.523}          & \multicolumn{1}{c|}{5.316}  & \multicolumn{1}{c|}{5.142} & \multicolumn{1}{c|}{5.496}     & \multicolumn{1}{c|}{3.648}  & \multicolumn{1}{c|}{2.693} & \multicolumn{1}{c|}{5.725}     & \multicolumn{1}{c|}{3.986}  & \multicolumn{1}{c|}{5.972} & 3.918     & \textbf{1.483} \\ \hline
\multicolumn{1}{l|}{\multirow{3}{*}{France}}    & Female                & \multicolumn{1}{c|}{2.247}     & 4.144     & \multicolumn{1}{c|}{\textbf{1.563}} & \multicolumn{1}{c|}{4.545}  & \multicolumn{1}{c|}{4.720} & \multicolumn{1}{c|}{3.832}     & \multicolumn{1}{c|}{3.384}  & \multicolumn{1}{c|}{3.459} & \multicolumn{1}{c|}{3.635}     & \multicolumn{1}{c|}{3.110}  & \multicolumn{1}{c|}{5.006} & 3.409     & 1.607          \\ \cline{2-15} 
\multicolumn{1}{l|}{}                           & Male                  & \multicolumn{1}{c|}{2.868}     & 5.276     & \multicolumn{1}{c|}{2.275}          & \multicolumn{1}{c|}{5.735}  & \multicolumn{1}{c|}{5.574} & \multicolumn{1}{c|}{6.114}     & \multicolumn{1}{c|}{4.378}  & \multicolumn{1}{c|}{4.267} & \multicolumn{1}{c|}{4.038}     & \multicolumn{1}{c|}{5.011}  & \multicolumn{1}{c|}{6.419} & 5.284     & \textbf{2.220} \\ \cline{2-15} 
\multicolumn{1}{l|}{}                           & Total                 & \multicolumn{1}{c|}{2.258}     & 4.364     & \multicolumn{1}{c|}{1.575}          & \multicolumn{1}{c|}{4.962}  & \multicolumn{1}{c|}{5.048} & \multicolumn{1}{c|}{4.628}     & \multicolumn{1}{c|}{3.782}  & \multicolumn{1}{c|}{3.666} & \multicolumn{1}{c|}{2.842}     & \multicolumn{1}{c|}{3.657}  & \multicolumn{1}{c|}{4.749} & 3.723     & \textbf{1.346} \\ \hline
\multicolumn{1}{l|}{\multirow{3}{*}{Japan}}     & Female                & \multicolumn{1}{c|}{6.870}     & 5.256     & \multicolumn{1}{c|}{2.418}          & \multicolumn{1}{c|}{3.087}  & \multicolumn{1}{c|}{3.421} & \multicolumn{1}{c|}{2.139}     & \multicolumn{1}{c|}{2.679}  & \multicolumn{1}{c|}{3.451} & \multicolumn{1}{c|}{4.666}     & \multicolumn{1}{c|}{2.740}  & \multicolumn{1}{c|}{4.800} & 2.281     & \textbf{1.920} \\ \cline{2-15} 
\multicolumn{1}{l|}{}                           & Male                  & \multicolumn{1}{c|}{3.419}     & 4.732     & \multicolumn{1}{c|}{\textbf{1.587}} & \multicolumn{1}{c|}{4.344}  & \multicolumn{1}{c|}{4.552} & \multicolumn{1}{c|}{3.863}     & \multicolumn{1}{c|}{3.436}  & \multicolumn{1}{c|}{3.225} & \multicolumn{1}{c|}{4.696}     & \multicolumn{1}{c|}{3.385}  & \multicolumn{1}{c|}{4.555} & 3.960     & 2.026          \\ \cline{2-15} 
\multicolumn{1}{l|}{}                           & Total                 & \multicolumn{1}{c|}{4.367}     & 4.435     & \multicolumn{1}{c|}{1.964}          & \multicolumn{1}{c|}{3.831}  & \multicolumn{1}{c|}{3.755} & \multicolumn{1}{c|}{2.541}     & \multicolumn{1}{c|}{2.887}  & \multicolumn{1}{c|}{3.624} & \multicolumn{1}{c|}{3.308}     & \multicolumn{1}{c|}{3.285}  & \multicolumn{1}{c|}{4.479} & 3.873     & \textbf{1.575} \\ \hline
\multicolumn{1}{l|}{\multirow{3}{*}{Portugal}}  & Female                & \multicolumn{1}{c|}{3.554}     & 5.391     & \multicolumn{1}{c|}{\textbf{2.957}} & \multicolumn{1}{c|}{7.348}  & \multicolumn{1}{c|}{7.395} & \multicolumn{1}{c|}{7.532}     & \multicolumn{1}{c|}{5.799}  & \multicolumn{1}{c|}{4.914} & \multicolumn{1}{c|}{5.263}     & \multicolumn{1}{c|}{6.132}  & \multicolumn{1}{c|}{8.490} & 5.131     & 3.125          \\ \cline{2-15} 
\multicolumn{1}{l|}{}                           & Male                  & \multicolumn{1}{c|}{5.575}     & 5.704     & \multicolumn{1}{c|}{3.161}          & \multicolumn{1}{c|}{7.458}  & \multicolumn{1}{c|}{7.627} & \multicolumn{1}{c|}{7.776}     & \multicolumn{1}{c|}{6.238}  & \multicolumn{1}{c|}{5.472} & \multicolumn{1}{c|}{9.321}     & \multicolumn{1}{c|}{7.414}  & \multicolumn{1}{c|}{9.147} & 6.517     & \textbf{3.038} \\ \cline{2-15} 
\multicolumn{1}{l|}{}                           & Total                 & \multicolumn{1}{c|}{4.134}     & 5.173     & \multicolumn{1}{c|}{2.598}          & \multicolumn{1}{c|}{7.459}  & \multicolumn{1}{c|}{7.480} & \multicolumn{1}{c|}{7.202}     & \multicolumn{1}{c|}{5.939}  & \multicolumn{1}{c|}{6.035} & \multicolumn{1}{c|}{5.879}     & \multicolumn{1}{c|}{6.381}  & \multicolumn{1}{c|}{9.143} & 6.234     & \textbf{2.134} \\ \hline
\end{tabular}
\end{adjustbox}
\label{tab:all_models}
\end{table}

\begin{table}[h!]
\centering
\caption{Mean of MAPE (\%), mean rank, std of the hybrid system ARIMA-LSTM with the Recursive approach compared to multivariate, statistical, and machine learning models. Best results in bold.}
 \begin{adjustbox}{max width=\textwidth}
\begin{tabular}
{ll|cc|cccccccccc|c}\hline
\multicolumn{2}{l|}{\multirow{3}{*}{Statistic}} & \multicolumn{2}{c|}{Multivariate}          & \multicolumn{10}{c|}{Single}                                                                                                                                     & Hybrid     \\ \cline{3-15} 
\multicolumn{2}{l|}{}                           & \multicolumn{1}{c|}{LC}        & PLAT      & \multicolumn{1}{c|}{ARIMA}     & \multicolumn{3}{c|}{LSTM}                       & \multicolumn{3}{c|}{MLP}                        & \multicolumn{3}{c|}{NBEATS} & ARIMA-LSTM \\ \cline{3-15} 
\multicolumn{2}{l|}{}                           & \multicolumn{1}{c|}{Recursive} & Recursive & \multicolumn{1}{c|}{Recursive} & Direct & MIMO  & \multicolumn{1}{c|}{Recursive} & Direct & MIMO  & \multicolumn{1}{c|}{Recursive} & Direct  & MIMO  & Recursive & Recursive  \\ \hline
\multicolumn{2}{l|}{Mean}                       & \multicolumn{1}{c|}{3.566}     & 4.589     & \multicolumn{1}{c|}{1.994}     & 4.990  & 4.991 & \multicolumn{1}{c|}{4.653}     & 3.833  & 3.677 & \multicolumn{1}{c|}{4.615}     & 4.157   & 5.823 & 6.676     & 1.905      \\
\multicolumn{2}{l|}{Mean rank}                  & \multicolumn{1}{c|}{3}         & 7         & \multicolumn{1}{c|}{2}         & 10     & 11    & \multicolumn{1}{c|}{9}         & 5      & 4     & \multicolumn{1}{c|}{8}         & 6       & 12    & 13        & 1          \\
\multicolumn{2}{l|}{Std}                        & \multicolumn{1}{c|}{1.327}     & 0.630     & \multicolumn{1}{c|}{0.558}     & 1.460  & 1.416 & \multicolumn{1}{c|}{1.883}     & 1.213  & 1.012 & \multicolumn{1}{c|}{1.726}     & 1.524   & 1.802 & 7.905     & 0.558      \\ \hline
\end{tabular}
\end{adjustbox}
\label{tab:mean_rank_std_all}
\end{table}

The mean MAPE and mean rank across all datasets place the ARIMA-LSTM model as the most accurate, reinforcing its robustness as a forecasting tool. The standard deviation (Std) of the hybrid model is equal to that of the ARIMA and is comparatively low, suggesting consistent and reliable performance across various datasets.

Table~\ref{tab:pd_all} shows the PD of the multi-step Recursive ARIMA-LSTM compared to the alternatives. The ARIMA model shows the smallest PD at 4.44\%, suggesting that although the ARIMA-LSTM model is more accurate, the difference is relatively modest. Classic multivariate models, LC and PLAT, displayed a more substantial PD of 46.58\% and 58.48\%, respectively, expressing a significant improvement in forecast precision when using ARIMA-LSTM.

In summary, compared with the alternatives, ARIMA-LSTM with the Recursive multi-step approach indicates a strong forecasting capability. Its hybrid structure, which can effectively capture the temporal dependencies of linear and nonlinear patterns, appears to be particularly suited to the complexities of mortality rate forecasting.

\begin{table}[h!]
\centering
\caption{Percentage difference of ARIMA-LSTM with Recursive multi-step approach in comparison with alternatives.}
\scalebox{0.95}{
\resizebox{4.5cm}{!}{\begin{tabular}{ll}
\hline
Model            & PD (\%) \\ \hline
ARIMA            & 4.44    \\
LC               & 46.58   \\
MLP-MIMO         & 48.18   \\
MLP-Direct       & 50.30   \\
N-BEATS-Direct    & 54.17   \\
PLAT             & 58.48   \\
MLP-Recursive    & 58.72   \\
LSTM-Recursive   & 59.05   \\
LSTM-Direct      & 61.82   \\
LSTM-MIMO        & 61.83   \\
N-BEATS-MIMO      & 67.28   \\
N-BEATS-Recursive & 71.46   \\ \hline
\end{tabular}}
}
\label{tab:pd_all}
\end{table}

\section{Conclusion}\label{sec:Conc}

The development of hybrid systems in forecasting has been highlighted because of their ability to model the linear and nonlinear patterns present in real-world time series. To evaluate the hybrid systems in a multi-step ahead mortality rate forecasting exercise, this study demonstrated the impact on accuracy for Recursive, Direct, and MIMO approaches and nonlinear ML forecasting models, such as MLP, LSTM, and N-BEATS, across 12 datasets. The study also compared the hybrid systems with a range of single and demographic forecasting models in the literature, utilizing the MAPE as the evaluation metric. 

Our findings revealed that the recursive approach was superior in modeling residual series of mortality rates, which in this case constituted short time series with potential nonlinear intricacies. 
Furthermore, the LSTM model was the most promising for residual series modeling, outperforming the alternatives in most scenarios. In addition, the best hybrid model with a recursive multi-step approach, that is, ARIMA-LSTM, showed the best forecast accuracy compared to other single time series models. Consequently, this study suggests that the most straightforward multi-step approach (i.e., recursive) should be employed for forecasting small datasets. Therefore, the results support the superiority of hybrid systems in the examined datasets. As points for improvement of the current work, the implementation of dynamic model selection approaches for residual modeling~\citep{de2021hybrid}, nonlinear combinations of methods~\citep{demattos:2017}, and uncertainty quantification should be considered.

\bibliographystyle{unsrtnat}
\bibliography{references}  






\end{document}